# Green Machine Learning via Augmented Gaussian Processes and Multi-Information Source Optimization

Antonio Candelieri[1*], Riccardo Perego[2], Francesco Archetti[2]

[1] Department of Economics, Management and Statistics,
University of Milano-Bicocca,
Milan, Italy

[2] Department of Computer Science, Systems and Communication,
University of Milano-Bicocca,
Milan, Italy

antonio.candelieri@unimib.it, riccardo.perego@unimib.it,
francesco.archetti@unimib.it

**Abstract.** Searching for accurate Machine and Deep Learning models is a computationally expensive and awfully energivorous process. A strategy which has been gaining recently importance to drastically reduce computational time and energy consumed is to exploit the availability of different information sources, with different computational costs and different "fidelity", typically smaller portions of a large dataset. The multi-source optimization strategy fits into the scheme of Gaussian Process based Bayesian Optimization. An Augmented Gaussian Process method exploiting multiple information sources (namely, AGP-MISO) is proposed. The Augmented Gaussian Process is trained using only "reliable" information among available sources. A novel acquisition function is defined according to the Augmented Gaussian Process. Computational results are reported related to the optimization of the hyperparameters of a Support Vector Machine (SVM) classifier using two sources: a large dataset – the most expensive one – and a smaller portion of it. A comparison with a traditional Bayesian Optimization approach to optimize the hyperparameters of the SVM classifier on the large dataset only is reported.

**Keywords:** Green AI, Green Machine Learning, Multi Information Source Optimization, Bayesian Optimization, Gaussian Processes.

# 1    Introduction

## 1.1    The Green AI challenge

Machine Learning (ML) models are computationally hungry: this is particularly true in the case of Deep Neural Networks (DNNs) in fields like computer vision (Bianco et al. 2020) and Natural Language Processing (NLP) (Kulkarni and Shivananda 2019): an approximate quantification of the financial and environmental costs for training and validating some of the neural network models in the NLP domain is reported in (Strumbell et al. 2019) and (Hao 2019) showing the amazing amount of energy consumed for training and validating a neural network model for NLP, which can generate the emission of an amount of carbon dioxide approximately five times the lifetime emissions of an average American car. No surprise that Green Machine Learning (Green-ML) and Green Artificial Intelligence (Green AI) (Schwartz et al. 2019; Yang et al. 2020) have recently emerged as new research topics.

This paper is focused on the issue of hyperparameter optimization (HPO), where *hyperparameters* are all the parameters of a model which are not updated during the learning and are used to configure either the model (e.g., number of layers of a deep neural network, etc.) or characterize the algorithm used in the training phase (learning rate for gradient descent algorithm, etc.) and even to include the choice of optimization algorithm itself and also the data features which are fed into the ML model.

HPO can be regarded as an *optimization outer loop* on top of ML model learning (*inner loop*) to find the set of hyperparameters leading to the lowest error on a validation set. This 2-tier optimization structure has several implications. First, the evaluation of the objective function of the outer loop is very expensive, as it requires learning a model and evaluating its performance on a validation set. This is usually repeated $k$ times in a $k$ fold-cross validation procedure. Moreover, the objective function is unknown and can only be observed pointwise with typically noisy evaluations. Secondly, the average value of the loss function does not reflect the true distribution of the data (which leads to the generalization error) and due to the relatively small size of the validation set, the variance of the average estimate obtained by cross validation can be high. Ignoring this uncertainty can result in sub-optimal configuration of hyperparameters. One must also take into account that the performance of the model is evaluated with some error, and thus finding the true optimum with a high precision is usually not critical: this fits nicely into in the Bayesian Optimization (BO) framework that is very sample efficient and yields an acceptable solution with relatively few function evaluations.

The outer loop optimization algorithm can be passive, like grid or pure random search, or "educated" to learn, from previous evaluations, the structure of the objective function, and to actively search where most interesting solutions are. Indeed, BO is a framework to model the learning process and to yield a principled quantification of uncertainty (Frazier 2018; Candelieri and Archetti 2019). BO has become the main approach to handle all the relevant steps in finding an accurate ML model: *Algorithm selection*, *Hyperparameter Optimization*, both recently integrated in the more general setting named CASH: *Combining Algorithm Selection and Hyperparameter optimization* (Kotthoff et al. 2017). This led to the definition of Automated Machine/Deep Learning (AutoML/AutoDL) (Hutter et al. 2019) and Neural Architecture Search (NAS) (Hutter et al. 2019; Lindauer and Hutter 2019), showing that different algorithms and values of its hyperparameters can result in significantly different performances (Wolpert et al. 2002; Melis et al. 2017).

Although the *active learning* inherent in BO and the ensuing sample efficiency are usually associated with the search for the best algorithm and its configuration (Shahriari et al. 2016), in terms of accuracy, they translate into significant cost and energy savings. For instance, the BERT (Bidirectional Encoder Representation from Transformer) model, now available in the Google Cloud, aimed at contextual representation in NLP, can require 4 days training sessions (with 110 million of DNN's parameters to be learned) (Strubell et al. 2019) which makes the NAS performed in the outer loop awfully expensive. Sample efficiency requires some assumption on the objective function and a model of learning from observations.

Probabilistic models commonly used in BO are Gaussian Processes (GPs) (Williams and Rasmussen 2006) and Random Forests (RFs) (Ho 1995) (here we do not discuss their relative merits in different problem classes). GPs are a powerful framework for reasoning about an unknown function $f$ given partial knowledge of its behavior obtained through function evaluations. GP leverages a principled estimate of predictive uncertainty towards a careful balance of *exploration* (increasing one's knowledge about $f$) and *exploitation* (focusing on the best points found so far).

The global hyperparameter optimization problem is usually defined as:

$$\min_{x \in X \subset \mathbb{R}^d} f(x) \tag{1}$$

where the search space $X$ is generally box-bounded, $f$ is the loss function and $x$ the values of the hyperparameters. We remark that $f$ is analytically unknown (also called *latent*) and only pointwise, usually noisy, evaluations can be obtained by querying it. We refer to this situation as black-box optimization.

BO leverages the fact that conditioning the GP on previous observations provides versatile regressors of the objective function. BO starts from a GP prior over $f$, encoded with parametric mean and kernel. The available observations are used to build the posterior distribution which is used to determine the learning policy, balancing exploration (high GP variance) and exploitation (low mean value).

Given the cost of evaluating the objective function, trial-&-error methods like random or grid search are not useful. Compared to a simple grid search, BO can identify a better solution for HPO, given the same number of configurations to evaluate. Given its modelling flexibility, BO can build a relatively cheap probabilistic surrogate of $f$, take advantage of related tasks (Swersky et al. 2013) or use problem specific priors (De Ath et al. 2020).

The strategy we follow here is to mitigate the high cost of hyperparameter optimization enabling the BO algorithm to trade-off the value of information gained from the evaluation of a hyperparameter configuration against its cost. In (Swersky et al. 2013) and (Klein et al. 2017) BO is used to evaluate models trained on randomly chosen subsets of data to obtain more, but less informative, evaluations. Two strategies aiming at the same target, which we do not consider here, are *curriculum learning*, which leverages a data-centric view training the model on increasingly larger datasets, and *continuation learning*, which leverages a model-centric view building a sequence of loss functions $L_1 \ldots L_r$, in which each $L_{i+1}$ is more difficult to optimize than $L_i$ and one can view each $L_i$ as a regularized version of $L_{i+1}$. (Aggarwal 2018),

These approaches could be interpreted as optimization problems in which multiple information sources are available, with every source approximating the actual black-box and expensive (loss) function, with a different cost for querying each information source. This setting is known as Multi-Information Source Optimization (MISO), or multi-fidelity optimization in the special case that the "fidelity" of each source is known a priori and independent on the value of the hyperparameters.

### 1.2 Multi Information Sources Optimization: Related works.

This problem was initially studied under the name of multi-fidelity optimization in which rather than a single objective $f$, we have a collection of information sources denoted with $f_1(x), \ldots, f_S(x)$. Each source has its own cost, $c_1, \ldots, c_S$, where $c_s > 0 \ \forall s = 1, \ldots, S - 1$, which controls the fidelity with lower $s$ giving higher fidelity: increasing the fidelity gives a more accurate estimate but at a higher cost. In the case of cross-validation the fidelity can be related to the number of iterations of the learning algorithm, the amount of data used in the training or the number of folds in the cross validation. In MISO the goal is to solve (1) while reducing the overall cost along the optimization process. MISO requires specific approaches to choose both the next location and source to evaluate, leading to a sequence $\{(s^{(1)}, x^{(1)}), \ldots, (s^{(N)}, x^{(N)})\}$. It is always possible to sort sources such that $c_s > c_{s+1}$; in the case that also $f(x)$ can be queried, then it is the most expensive source, so we can set $f(x) = f_1(x)$ without loss of generality.

In the early work about multi-fidelity $f_s(x)$ were assumed to be ordered in terms of accuracy and cost: in more general problems of multi-information source optimization we only assume the function $f(x)$ taking a design input $x$, the objective and $f_s(x)$ being the sources with different biases, different amount of noise and different costs.

MISO has been gaining increasing attention in the last years, also beyond ML. An example in engineering design is the finite element method, where models with cost and fidelity can be obtained using different mesh values. Cheap approximations do not represent accurately the optimization targets, but still can offer an indication of the sensitivity of the output to changes in the parameters. Also, output data from physical prototypes can be integrated in the optimization framework as an additional information source, with fidelity depending on the application and the experimental setting. The application domain which has first exploited the advantages offered by multi-fidelity and multi-information source optimization is aerodynamics: in (Chaudhuri et al. 2019) and (Lam et al. 2015) is presented an approach that adaptively updates a multi-fidelity surrogate on multiple information sources and without any assumption about hierarchical relations among them.

In a seminal paper (Swersky et al. 2013) the use of small datasets to quickly optimize the hyperparameters of a ML model for large datasets has been proposed. The method shows that it is possible to transfer the knowledge gained from previous optimizations to new tasks in order to speed up k-fold cross validation. The algorithm dynamically chooses which dataset to query in order to yield the most information per unit cost. In (Kandasamy et al. 2016) a multi-fidelity bandit optimisation based on Gaussian Process (GP) approximations of all the sources is proposed. The algorithm is named Multi-Fidelity Gaussian Process Upper Confidence Bound (MF-GP-UCB) and resulted able to explore the search space – spanned by the hyperparameters of the ML algorithm to optimize – using the lower fidelity sources and then exploit the higher fidelity sources in successively smaller regions, converging to the optimum. FABOLAS (FAst Bayesian Optimization on LArge dataSets) (Klein et al. 2017) is an approach for HPO on large datasets: at each iteration, it selects an hyperparameters configuration and a dataset size to use for optimizing hyperparameters for the entire dataset. Results are reported for HPO of Support Vector Machines (SVM) and DNNs, with FABOLAS often providing good solutions significantly faster than "vanilla" BO-based HPO on the full dataset. The approach in (Poloczek et al. 2017) uses a GP with a kernel working on a space consisting of both the search space (spanned by the hyperparameters to optimize) and the information sources. In (Ghoreishi and Allaire 2019) an approach incorporating correlations both within and among information sources is proposed. This allows to exploit the information collected over all the sources and then fusing them in a unique fused GP. Furthermore, the constrained setting is considered, where also constraints can be queried on multiple information sources.

A different approach has been proposed in (Ariafar et al. 2020) Importance based Bayesian Optimization (IBO) which models a distribution over the location of optimal hyperparameter configuration and allocates experimental budget according to cost adjusted expected reduction in entropy (Hennig and Schuler 2012). Higher fidelity observations provide a larger reduction in entropy, albeit at a higher evaluation cost.

To properly quantify predictive uncertainty, it is important for a learning system to recognize different types of uncertainty arising in the modelling process (Liu et al. 2019). Two types of uncertainty must be considered: aleatoric and epistemic. Aleatoric arises due to the stochastic variability of the data generating process, imperfect sensors, epistemic arises due to our lack of knowledge about the data generating mechanism. A model epistemic uncertainty can be reduced by collecting more data and takes two forms: parametric uncertainty, that is uncertainty associated with estimating the model parameters under the current model specification and structural uncertainty that reflects the measure in which a model is sufficient to describe the data ,i.e. whether there exists a systematic discrepancy.

### 1.3    Our Contributions

The main contributions of this paper can be summarized as follows:

- A new GP called augmented GP which does not require a kernel working in the $x, s$ space of hyperparameters and sources. Relations among sources are captured by a simplified and computationally cheap discrepancy measure (related to the epistemic error), used to select "reliable" evaluations to fit the proposed GP and included into a new acquisition function
- A new acquisition function based on U/LCB but implementing a sparsification strategy. Indeed, the proposed GP results *sparse*, reducing the computational cost for fitting it (i.e., the number of evaluations raised power of three).
- The new GP mitigates the computational problems in estimating nonparametric regression which is inherently difficult in high dimensions with known lower bounds depending exponentially on dimension.
- Making MISO energy-efficient itself by selecting a subset of "reliable" evaluations among all those performed over all the sources. Only this subset is used to fit a GP differently from the fused GP in (Ghoreishi and Allaire 2019).
- Demonstrating, empirically, the benefit provided by our approach on an HPO task aimed at optimally tuning a Support Vector Machine classifier on a large dataset.

## 2    Background

### 2.1    Gaussian Processes

The global optimization problem is defined as:

$$\min_{x \in X \subset \mathbb{R}^d} f(x) \qquad (1)$$

where the search space $X$ is generally box-bounded, no analytical expression is known of $f(x)$, whose shape can only be learned from its evaluations.

One way to interpret a Gaussian Process (GP) regression model is to think of $f$ as a latent function defining a distribution over functions, and with inference taking place directly in the space of functions (i.e., *function-space view*) (Williams and Rasmussen, 2006). A GP is a collection of random variables, any finite number of which have a joint Gaussian distribution. A GP is completely specified by its mean function $\mu(x)$ and covariance function $cov\big(f(x), f(x')\big) = k(x, x')$:

$$\mu(x) = \mathbb{E}[f(x)] \tag{2}$$

$$cov\big(f(x), f(x')\big) = k(x, x') = \mathbb{E}[(f(x) - \mu(x))(f(x') - \mu(x'))]$$

and will write the GP as:

$$f(x) \sim GP\big(\mu(x), k(x, x')\big) \tag{3}$$

Usually, for notational simplicity we will take the prior of the mean function to be zero, although this is not necessary.

A simple example of a GP can be obtained from a Bayesian linear regression model $f(x) = \phi(x)^T w$ with prior $w = \mathcal{N}(0, \Sigma_p)$, where $\phi(x)$ and $w$ are $p$-dimensional vectors. More precisely $\phi(x)$ is a function mapping the $d$-dimensional vector $x$ into a $p$-dimensional vector.

Thus, the equations for mean and covariance become:

$$\mathbb{E}[f(x)] = \phi(x)^T \mathbb{E}[w] = 0 \tag{4}$$

$$\mathbb{E}[f(x)f(x')] = \phi(x)^T \mathbb{E}[ww^T]\phi(x') = \phi(x)^T \Sigma_p \phi(x')$$

This means that $f(x)$ and $f(x')$ are jointly Gaussian with zero mean and covariance given by $\phi(x)^T \Sigma_p \phi(x')$.

As consequence, the function values $f(x_1), \ldots, f(x_n)$ obtained at $n$ different points $x_1, \ldots, x_n$, are jointly Gaussian. The covariance function assumes a critical role into the GP modelling, as it specifies the distribution over functions. To see this, we can draw samples from the distribution of functions evaluated at any number of points; in detail, we choose a set of input points $X_{1:n} = (x_1, \ldots, x_n)^T$ and then compute the corresponding covariance matrix elementwise. This operation is usually performed by using predefined covariance functions allowing to write covariance between *outputs* as a function of *inputs* (i.e., $cov\big(f(x), f(x')\big) = k(x, x')$). Finally, we can generate a random Gaussian vector as:

$$f(X_{1:n}) \sim \mathcal{N}\big(\mathbf{0}, \mathrm{K}(X_{1:n}, X_{1:n})\big) \tag{5}$$

and plot the generated values as a function of the inputs. This is basically known as *sampling from prior*.

Let $\mathbf{X}_{1:n} = \left\{ x^{(1)}, \dots, x^{(n)} \right\}$ denotes a set of $n$ locations into the search space $\mathcal{X}$ and $\mathbf{y} = \left\{ y^{(1)}, \dots, y^{(n)} \right\}$ the associated function values, with $y^{(i)} = f\left( x^{(i)} \right)$ or, in the noisy setting, $y^{(i)} = f\left( x^{(i)} \right) + \varepsilon$ with $\varepsilon \sim \mathcal{N}(0, \lambda^2)$. Then, the GP's mean and variance are conditioned as follows:

$$\mu(x) = \mathbf{k}(x, \mathbf{X}_{1:n})[\mathbf{K} + \lambda^2 \mathbf{I}]^{-1} \mathbf{y} \qquad (6)$$

$$\sigma^2(x) = k(x, x) - \mathbf{k}(x, \mathbf{X}_{1:n})[\mathbf{K} + \lambda^2 \mathbf{I}]^{-1} \mathbf{k}(\mathbf{X}_{1:n}, x) \qquad (7)$$

with $k$ a kernel function, $\mathbf{k}(x, \mathbf{X}_{1:n})$ a vector whose $i$th component is $k\left( x, x^{(i)} \right)$ and $\mathbf{K}$ an $n \times n$ matrix with entries $\mathbf{K}_{ij} = k\left( x^{(i)}, x^{(j)} \right)$. Finally, $\mathbf{k}(\mathbf{X}_{1:n}, x)$ is the transposed version of $\mathbf{k}(x, \mathbf{X}_{1:n})$.

Following, a simple example of 5 different samples drawn at random from a GP prior and posterior, respectively. The posterior is conditioned on 6 function observations.

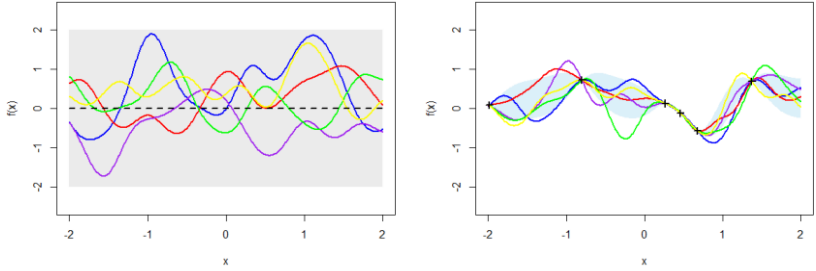

**Fig. 1 Sampling from prior vs sampling from posterior (for the sake of simplicity, we consider the noisy-free setting)**

It is easy to notice that the mean prediction is a linear combination of $n$ functions, each one centred on an evaluated point. This allows to write $\mu(x)$ as:

$$\mu(x) = \sum_{i=1}^{n} \alpha_i k(x, x_i) \qquad (8)$$

where the vector $\alpha = [\mathrm{K}(X_{1:n}, X_{1:n}) + \lambda^2 I]^{-1}\mathrm{y}$ and $\alpha_i$ is the $i$-th component of the vector $\alpha$, given by the product between the $i$-th row of the matrix $[\mathrm{K}(X_{1:n}, X_{1:n}) + \lambda^2 I]^{-1}$ and the vector $\mathrm{y}$.

This means that, to make a prediction at a given $x$, we only need to consider the $(n + 1)$-dimensional distribution defined by the $n$ function evaluations performed so far and the new point $x$ to evaluate.

The values of the hyperparameters are usually unknown a priori and are set up depending on the observations $D_{1:n}$, usually, via Marginal Likelihood maximization.

GP's hyperparameters $\gamma$ appear non-linearly in the kernel matrix $K$ and a closed-form solution to maximizing the marginal likelihood cannot be found in general. In practice, gradient-based optimization algorithms are adopted to find a (local) optimum of the marginal likelihood (e.g., conjugate gradients or BFGS).

An interesting generalization has been recently proposed in (Berkenkamp et al., 2019) where the values of hyperparameters are modified by iteratively reducing the characteristic length-scale instead of setting them up through Marginal Likelihood maximization.

## 2.2   Kernels

A kernel function (aka covariance function) is the crucial ingredient in a GP predictor, as it encodes assumptions about the function to approximate. it is clear that the notion of similarity between data points is crucial; it is a basic assumption that points which are close are likely to have similar target values $y$, and thus function evaluations that are near to a given point should be informative about the prediction at that point. Under the GP view it is the covariance function that defines nearness or similarity.

*Squared Exponential (SE) kernel:*

$$k_{SE}(x, x') = e^{-\frac{\|x - x'\|^2}{2\ell^2}}$$

With $\ell$ known as *characteristic length-scale*. A large value of the length-scale will map $x$ to a narrower range of values, while a small length-scale does the opposite. Consequently, a large length-scale implies long-range correlations, whereas a short length-scale makes function values strongly correlated only if their respective inputs are very close to each other. This kernel is infinitely differentiable, meaning that the sample paths of the corresponding GP are very "smooth".

Another way to look at l is through the expected number of 0-upcrossings which is proportional to $1/\ell$. Then $\ell$ is proportional to the expected length before crossing 0, hence the name length scale.

SE is the most widely used kernel because it is easy to code, relatively robust to misspecification and guarantees a positive definite covariance regardless of input dimensions. One must anyway bear in mind that it's particularly liable to numerical ill conditioning of the kernel matrix.

*Matérn kernels:*

$$k_{Mat}(x, x') = \frac{2^{1-\nu}}{\Gamma(\nu)} \left( \frac{|x - x'|\sqrt{2\nu}}{\ell} \right)^{\nu} K_{\nu} \left( \frac{|x - x'|\sqrt{2\nu}}{\ell} \right)$$

With two hyperparameters $\nu$ and $\ell$, and where $K_\nu$ is a modified Bessel function. Note that for $\nu \to \infty$ we obtain the SE kernel.

The Matérn covariance functions become especially simple when $\nu$ is half-integer: $\nu = p + 1/2$, where $p$ is a non-negative integer.

. The formula can be rewritten as the product of an exponential and polynomial terms of order $p - 1$.

The advantages of the simplified covariance Matérn function is that there are no Bessel functions, no sum of factorials nor fraction of gammas as reported in (Gramacy et al. 2020). This is important because *the evaluation of the Bessel function can be as computationally demanding as the matrix inversion*

The most widely adopted versions, specifically in the Machine Learning community, are $\nu = 3/2$ and $\nu = 5/2$.

$$k_{\nu=3/2}(x,x') = \left(1 + \frac{|x-x'|\sqrt{3}}{\ell}\right) e^{-\frac{|x-x'|\sqrt{3}}{\ell}}$$

$$k_{\nu=5/2}(x,x') = \left(1 + \frac{|x-x'|\sqrt{5}}{\ell} + \frac{(x-x')^2}{3\ell^2}\right) e^{-\frac{|x-x'|\sqrt{5}}{\ell}}$$

Choosing $p=0$ one obtains the exponential family, $p=0$ implies $\nu=1/2$ which is appropriate for rough surfaces.

Sample path of latent $f$ under a GP with Matérn will be $k$-times differentiable iff $\nu$ larger than $k$.

One great advantage of Matérn is that at least for small $\nu$ it creates covariance matrices that are better conditioned than SE.

The exponential kernel is also called the Laplace kernel and has a strong link with Mondrian kernels which results in Gaussian models conceptually close to Random Forests (Lévesque et al. 2017).

*Rational Quadratic Covariance Function*

$$k_{RQ}(x,x') = \left(1 + \frac{(x-x')^2}{2\alpha\ell^2}\right)^{-\alpha}$$

where $\alpha$ and $\ell$ are two hyperparameters. This kernel can be considered as an infinite sum (*scale mixture*) of SE kernels, with different characteristic length-scales.

The afore mentioned kernels are just the most widely adopted in GP regression.

More details and a most comprehensive set of covariance functions are reported in (Williams and Rasmussen, 2006), and (Gramacy 2020) including non-stationary kernels and dot product kernels.

Some issues on kernel have been considered in recent publications, such as: kernel composition, safe optimization in relation to cognition (Schultz et al., 2018) as well as kernel learning, adaptation and sparsity in order to deal with functions that are smooth in a subset of their domain and can vary rapidly in another as analysed in (Peifer et al., 2019) from the viewpoint of computational complexity in the framework of RKHS (Reproducing Kernel Hilbert Spaces).

A space-temporal kernel has been proposed in (Nyikosa et al., 2018) to allow the GP to capture all the instances of the function over time and track a temporally evolving minimum.

## 2.3 Acquisition functions

The acquisition function is the mechanism to implement the trade-off between *exploration* and *exploitation* in BO. More precisely, any acquisition function aims to guide the search of the optimum towards points with potential low values of objective function either because the prediction of $f(x)$, based on the probabilistic surrogate model, is low or the uncertainty is high (or both). Indeed, *exploiting* means to target the area providing more chance to improve the current solution (with respect to the current surrogate model), while *exploring* means to move towards less explored regions of the search space where predictions based on the surrogate model have a higher variance.

### 2.3.1 Probability of Improvement

*Probability of Improvement* (PI) was the first acquisition function proposed in the literature (Kushner, 1964):

$$PI(x) = P(f(x) \leq f(x^+) + \xi) = \mathbf{\Phi}\left(\frac{f(x^+) - \mu(x) - \xi}{\sigma(x)}\right)$$

where $f(x^+)$ is the best value of the objective function observed so far, $\mu(x)$ and $\sigma(x)$ are mean and standard deviation provided by (6) and (7), and $\mathbf{\Phi}(\cdot)$ is the normal cumulative distribution function. . The parameter $\xi$ is introduced to modulates the balance between exploration and exploitation. More precisely, $\xi = 0$ is towards exploitation while $\xi > 0$ is more towards exploration.

The next point to evaluate is chosen according to: $x_{n+1} = \underset{x \in X}{\mathrm{argmax}}\, PI(x)$

### 2.3.2 Expected Improvement

*Expected Improvement* (EI), proposed initially proposed in (Močkus et al., 1975) and then made popular in (Jones et al., 1998) , measures the expectation of the improvement on $f(x)$ with respect to the predictive distribution of the probabilistic surrogate model.

$$EI(x) = \begin{cases} (f(x^+) - \mu(x) - \xi)\mathbf{\Phi}(Z) + \sigma(x)\phi(Z) \; if \; \sigma(x) > 0 \\ 0 \; if \; \sigma(x) = 0 \end{cases}$$

$$Z = \begin{cases} \dfrac{f(x^+) - \mu(x) - \xi}{\sigma(x)} \ if \ \sigma(x) > 0 \\ \qquad\quad 0 \ if \ \sigma(x) = 0 \end{cases}$$

The parameter $\xi$ in order to actively manage the trade-off between exploration (larger values) and exploitation (smaller): $\xi$ should be adjusted dynamically to decrease monotonically with the function evaluations.

The next point to evaluate is chosen according to: $x_{n+1} = \underset{x \in X}{\text{argmax}} \, EI(x)$

### 2.3.3 Upper/Lower Confidence Bound

*Confidence Bound* – where Upper and Lower are used, respectively for maximization and minimization problems – is an acquisition function that manage exploration-exploitation by being optimistic in the face of uncertainty, in the sense of considering the best-case scenario for a given probability value (Auer, 2002).

For the case of minimization, LCB is given by:

$$\text{LCB}(x) = \mu(x) - \xi\sigma(x)$$

where $\xi \geq 0$ is the parameter to manage the trade-off between exploration and exploitation ($\xi = 0$ is for pure exploitation; on the contrary, higher values of $\xi$ emphasizes exploration by inflating the model uncertainty). For this acquisition function there are strong theoretical results, originated in the context of multi-armed bandit problems, on achieving the optimal regret derived by (Srinivas et al. 2012). For the candidate point $x_n$ we observe instantaneous regret $r_n = f(x_n) - f(x^*)$. The cumulative regret $R_N$ after $N$ function evaluations is the sum of instantaneous regrets: $R_N = \sum_{n=1}^{N} r_n$. A desirable asymptotic property of an algorithm is to be no-regret: $\lim_{N \to \infty} \frac{R_N}{N} = 0$. Bounds on the average regret $\frac{R_N}{N}$ translate , bounding $R_N$ by a quantity sublinear in $T$, to convergence rates: $f(x^+) = \min x_{n \leq N} \, f(x_n)$ in the first $N$ function evaluations is no further from $f(x^*)$ than the average regret. Therefore, $f(x^+) - f(x^*) \to 0$, with $N \to \infty$ and so a no regret algorithm will converge to a subset of the global minimizers.

A wide analysis of the convergence rate of $R_N$ in the case of Matérn kernel, for different values of $d$ and $\nu$, is given in (Vakili et al. 2020).

The following figure shows how the selected points changes depending on $\xi$.

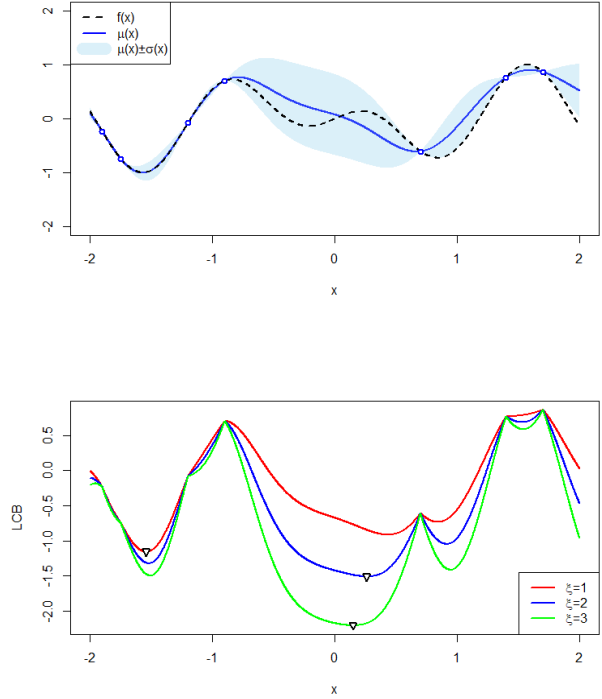

**Fig. 2. GP trained depending on 7 observations (top), LCB with respect to different values of $\xi$ and min values corresponding to the next point to evaluate (bottom).**

Finally, the next point to evaluate is chosen according to $x_{n+1} = \underset{x \in X}{\operatorname{argmin}} \, LCB(x)$, in the case of a minimization problem, or $x_{n+1} = \underset{x \in X}{\operatorname{argmax}} \, UCB(x)$ in the case of a maximization problem.

From the perspective of BO a particularly interesting bandit problem is the kernelized continuum armed bandit-problem (Srinivas et al. 2010). Here $f$ is assumed to be in the closure of functions on $X$ expressible as a linear combination of a feature embedding parametrised by a kernel $k$. The properties of the functions in the resulting space referred as the RKHS of $k$, are determined by the choice of the kernel. For a SE kernel the RKHS contains only infinitely differentiable functions. The Matérn kernel is parametrised by a smoothness parameter $\nu$, for a given $\nu$, the Matérn RKHS contains all functions $\nu$ times differentiable.

The optimization of the acquisition function leads to the next location to be queried, $x^{(n+1)}$, and, consequently, to a sequence of locations generated $\{x^{(1)}, \dots, x^{(N)}\}$ over the BO process, with $N$ the overall number of function evaluations at the end of the process.

In this paper we use Lower Confidence Bound, largely adopted in GP-based BO and with a convergence proof under an appropriate scheduling of the internal parameter $\beta^{(n)}$ (Srinivas et al. 2012) which balances between exploration and exploitation.

$$LCB^{(n)}(x) = \mu^{(n)}(x) - \sqrt{\beta^{(n)}}\sigma^{(n)} \tag{9}$$

where the apex related to the current iteration $n$ has been included to highlight that the value of $\beta$ changes over BO iterations, as well as the conditioned GP's mean and standard deviation. Confidence Bound has been successfully applied in MISO, such as in (Kandasamy et al. 2016). (Wilson et al. 2018) point out that the shape of the acquisition function may have large flat regions which, in particular in high dimensional spaces, make its optimization problematic and propose a Monte Carlo evaluation of acquisition function amenable to gradient-based optimization and identify a family of acquisition functions, including EI and UCB, whose characteristics allow to use of greedy approaches for their maximization.

A specific problem in MISO is related to the acquisition function: a direct translation of the popular expected improvement causes $EI = 0$ leading to querying only the highest fidelity source. According to (Poloczek et al. 2017) and (Ghoreishi et al. 2019) Knowledge Gradient, Entropy Search and Predictive Entropy Search can be applied. However, their computation and optimization are computationally more expensive: for this reason, in this paper we consider L/UCB and build on it a new acquisition functions specifically designed for MISO.

# 3 The proposed Multi Information Source Optimization - Augmented Gaussian Process (MISO-AGP)

## 3.1 Augmented GP

The MISO approach proposed in this paper is based on the idea of training a GP on a "reliable" subset of all the function evaluations performed so far over all the information sources. We refer to this GP as *Augmented Gaussian Process* (AGP) and consequently named our approach MISO-AGP. The term "augmented" is used to highlight that the set of function evaluations to train the AGP starts from those performed on the most expensive source and then it is "augmented" by selecting evaluations performed on some other source. Before explaining how the selection process is performed, we introduce some useful notations.

Let $D_s = \left\{\left(x^{(i)}, y_s^{(i)}\right)\right\}_{i=1,\dots,n_s}$ denotes the $n_s$ function evaluations performed so far on the source $s$. For each source $s$ a specific GP, $\mathcal{G}_s$, is trained on the current $D_s$. Let introduce a *model discrepancy* measure, $\eta(x, \mathcal{G}, \mathcal{G}')$, between two GPs. Differently from other papers, such as (Poloczek et al. 2017; Ghoreishi et al. 2019), we compute it simply as:

$$\eta(x, \mathcal{G}, \mathcal{G}') = |\mu(x) - \mu'(x)| \tag{10}$$

with $\mu(x)$ and $\mu'(x)$ the mean functions of the two GPs. It is also important to note that $\eta(x, \mathcal{G}, \mathcal{G}')$ depends on $x$. Indeed, in MISO we do not know a-priori the fidelity of each source and it could be not constant over $\mathcal{X}$.

Assume that $f(x)$ can be queried at the highest cost, that is $f(x) = f_1(x)$. Thus, the set of evaluations to train the AGP consists of $D_1$ "augmented" by:

$$\tilde{D} = \{(\tilde{x}, \tilde{y}) : \exists z : (\tilde{x}, \tilde{y}) \in D_z \ \wedge \ \eta(x, \mathcal{G}_1, \mathcal{G}_z) < m \ \sigma_1(x)\} \tag{11}$$

with $m$ a technical parameter of the MISO-AGP algorithm. We used $m = 1$ (i.e., around 68% of observations normally distributed are in the interval mean $\pm$ standard deviation). Thus, function evaluations on cheaper sources, having a discrepancy lower than the threshold given in (11), are considered "reliable" to be merged with those collected on the most expensive source. Let $\tilde{D}$ denotes the augmented set of function evaluations, such that $\bar{D} = D_1 \cup \tilde{D}$, the AGP $\hat{\mathcal{G}}$ is trained on $\bar{D}$, leading to $\hat{\mu}(x)$ and $\hat{\sigma}(x)$, computed according to (6-7). An example is reported in Fig. 1.

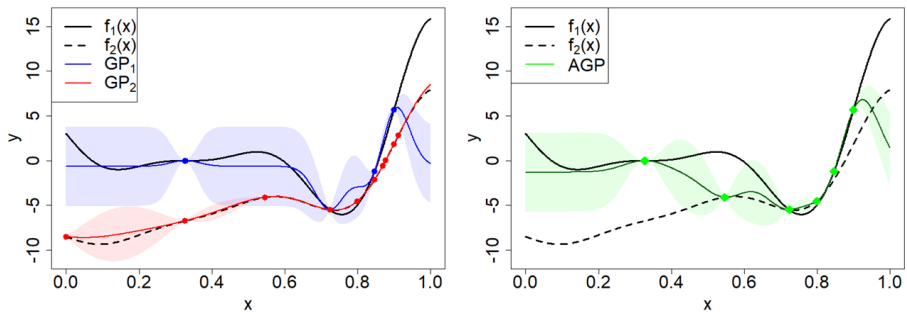

**Fig. 3.** An example of AGP on a 1-dimensional MISO minimization problem with two information sources. (Left) the two GPs trained on each source; (right) the AGP: only 3 evaluations on the cheaper source (around $x = 0.5$, $x = 0.7$ and $x = 0.8$) are selected to "augment" the evaluations on the expensive one. This reduces, at the same time, the uncertainty near the global minimum of $f_1$ and the number of evaluations for training the AGP (6 out of the 14 overall).

### 3.2 Acquisition function in MISO-AGP algorithm

Following the training of the AGP, an acquisition function must be used to choose the next pair *source-location* to query, that is $(s', x')$. We consider the framework of U/LCB:

$$(s', x') = \underset{\substack{x \in \mathcal{X} \subset \mathbb{R}^d \\ s=1,\dots,S}}{\operatorname{argmax}} \left\{ \frac{-\left( \hat{\mu}(x) - \sqrt{\beta^{(n)}} \hat{\sigma}(x) \right)}{c_s \left( 1 + \eta(x, \hat{\mathcal{G}}, \mathcal{G}_s) \right)} \right\} \quad (12)$$

where $n$ is the number of function evaluations into $\widehat{D}$. The numerator is the opposite of the AGP's LCB (indeed, we are maximizing (12)), penalized by the cost of the source $s$ and the model discrepancy between the AGP $\hat{\mathcal{G}}$ and $\mathcal{G}_s$, at the location $x$.

There is the chance that $x'$ could be too close to some previous function evaluations on $s'$. This behaviour arises when BO is converging to a (local/global) optimum and leads to a well-known instability issue in GP training, that is ill-conditioning in the inversion of the matrix $[\mathbf{K} + \lambda^2 \mathbf{I}]$. This instability issue occurs even more frequently and quickly in the noise-free setting (i.e., $\lambda = 0$). To avoid this undesired behaviour – leading to wasting evaluations without obtaining any improvement and/or risking occurring in the instability issue – we introduce the following correction.

Given $(s', x')$ from (12), if $\exists (x^{(i)}, y^{(i)}) \in D_{s'} \wedge \|x' - x^{(i)}\|^2 < \delta$

$$s' \leftarrow 1 \text{ and } x' = \underset{x \in \mathcal{X} \subset \mathbb{R}^d}{\operatorname{argmax}} \sigma_1(x) \quad (13)$$

with $\delta > 0$ the second MISO-AGP's technical parameter. In other words, we set the acceptable level of approximation, $\delta$, in locating the optimizer and, in the case that $x'$ is closer than $\delta$ to another evaluation on $s'$, then we prefer to "spend our budget" in reducing uncertainty on the most expensive source.

The MISO-AGP algorithm is summarized in the following.

---

**Algorithm: MISO-AGP**

**Input:**
    $f_1(x), \dots, f_S(x)$; $c_1, \dots, c_S$; $\mathcal{X}$; max cumulated cost $\bar{C}$; max iterations $N$;
    set $m$ and $\delta$ (MISO-AGP's technical parameters)

**Initialization:**
    $D_s = \left\{ \left( x^{(i)}, y_s^{(i)} \right) \right\}_{i=1,\dots,n_s} \forall s = 1, \dots, S$ and with $n_s$ initial evaluations on locations randomly sampled in $\mathcal{X}$

**Main:**
    $c \leftarrow 0;\ n \leftarrow 0;$
    **while** ($c < \bar{C}$ AND $n < N$ ) **do**
        train $\mathcal{G}_s$ on $D_s \forall s = 1, \dots, S \Rightarrow \mu_s(x), \sigma_s(x)$
        build $\widehat{D} = D_1 \cup \widetilde{D}$ with $\widetilde{D}$ defined in (11)
        train the AGP $\hat{\mathcal{G}}$ on $\widehat{D} \Rightarrow \hat{\mu}(x), \hat{\sigma}(x)$
        choose $(s', x')$ according to (12)
        **if** $\exists x^{(i)} : (x^{(i)}, y^{(i)}) \in D_{s'} \wedge \|x' - x^{(i)}\|^2 < \delta$ **then**
            $(s', x')$ according to (13)

**endif**
    query source $s'$ at location $x'$ and observe $y_{s'}$
    update $D_{s'} \leftarrow D_{s'} \cup \{(x', y_{s'})\}$
    $c \leftarrow c + c_{s'}$
    $n \leftarrow n + 1$
**endwhile**
**Output:**
    build $\tilde{D} = D_1 \cup \tilde{D}$ with $\tilde{D}$ defined in (11)
    return $(x^+, y^+) \in \tilde{D}$: $y^+ = \min\limits_{i=1,\dots,\tilde{n}} \{y^{(i)}\}$ with $\tilde{n}$ the function evaluations in $\tilde{D}$

## 4     Experimental setting

### 4.1    C-SVC with RBF kernel

To validate our MISO-AGP approach we designed an HPO task whose goal is to optimally and efficiently tune the hyperparameters of a Support Vector Machine (SVM) classifier on a large dataset. More precisely, we consider a C-SVC with a Radial Basis Function (RBF) kernel and the "MAGIC Gamma Telescope" dataset [1].

We chose C-SVC due to its relative inefficiency on large datasets: computational complexity for training a C-SVC, on a given hyperparameters configuration, is the number of instances raised the power of three. The C-SVC's hyperparameters to optimize are the regularization term, $C$, and $\gamma$ in the RBF kernel: $k_{RBF}(x, x') = e^{-\gamma \|x - x'\|^2}$.

The MAGIC dataset is generated by a Monte Carlo program (Heck et al. 1998), to simulate registration of high energy gamma particles in a ground-based atmospheric Cherenkov gamma telescope using the imaging technique. The overall dataset consists of 19'020 instances: 12'332 of the class "gamma (signal)" and 6'688 of the class "hadron (background)", with each instance represented by 10 continuous features. We have performed a pre-processing consisting in scaling all the dataset features in [0,1].

### 4.2    MISO-AGP setting

Following the notation used in this paper, MISO-AGP will be used to minimize $f(x)$, that is the misclassification error of a C-SVC, computed on 10-fold cross validation, on the MAGIC dataset. The search space $\mathcal{X}$ is 2-dimensional and box-bounded, spanned by the two C-SVC's hyperparameters $C \in [10^{-2}, 10^2]$ and $\gamma \in [10^{-4}, 10^4]$. We adopt a logarithmic scaling of the search space, a usual procedure suggested in AutoML for hyperparameters varying within ranges of this scale.

We have defined two different sources: the first provides the misclassification error obtained via 10-fold cross validation of a C-SVC configuration using the entire MAGIC

---

dataset (i.e., $f_1(x) = f(x)$). The second (i.e., $f_2(x)$) performs the same computation but using a smaller portion of the data (just 5% through stratified sampling).

Energy required to perform 10-fold cross validation is basically associated to the computational time, which we consider as a proxy for the sources' costs. Since computational time can also depend on the values of C-SVC's hyperparameters, we have run a sample of 10 hyperparameters configurations on both the two sources and used the average computational times for estimating reference values for $c_1$ and $c_2$. More precisely, computational time required by $f_1(x)$ is, on average, 320 times that required by $f_2(x)$. Thus, we set $c_2 = 1$ and, consequently, $c_1 = 320$.

The kernel used to model the covariance function, for all the GPs, including the AGP, is the Squared Exponential kernel, whose hyperparameters are set via Maximum Loglikelihood Estimation during the GP training. The acquisition function (12) and, in case, the correction (13) are both optimized via L-BFGS.

As initialization, 3 hyperparameters configurations are sampled in $\mathcal{X}$ via Latin Hypercube Sampling. Then, 30 further function evaluations are used by MISO-AGP to optimize over sources. We decided not to set a limit on the cumulated cost but to use this value to make considerations on the efficiency of the proposed approach with respect to BO applied only on the most expensive source. To mitigate the effect of initial randomness, 10 different runs of MISO-AGP and BO have been performed and compared: at each run, the two approaches share the same initialization.

As metrics, we consider the best function value observed so far. It is usually named "best seen" in BO and simply defined as $y_+^{(n)} = \min_{i=1,\dots,n} \{y^{(1)}, \dots, y^{(n)}\}$ – because we are considering the minimization of the misclassification error. However, this definition is no more valid in the case of the AGP. Suppose that, at a certain iteration, a function evaluation on a cheaper source is selected to fit the AGP and that corresponds to the best seen up to that iteration. At the next iteration, it could be not selected and, consequently, it cannot be considered as the best seen any longer. More formally, let $\hat{y}_+^{(n)}$ denotes the "augmented best seen", $\hat{y}_+^{(n)} = \min_{i=1,\dots,p} \{y^{(1)}, \dots, y^{(p)}\}$, with $p < n$ because only a subset of the evaluations on all the sources is used to train the AGP. In the case that $\hat{y}_+^{(n-1)} \notin \{y^{(1)}, \dots, y^{(p)}\} \Rightarrow \hat{y}_+^{(n)} \lesseqgtr \hat{y}_+^{(n-1)}$; in other terms, contrary to the common "best seen", the "augmented best seen" could not be monotone over the function evaluations.

## 5 Results

The following figure (Fig. 2) summarizes the results of the study. The best value of the misclassification error is reported with respect to the cost cumulated over the MISO-AGP and BO iterations, separately. Solid lines represent the mean over the 10 independent runs, while shaded areas represent the standard deviations. As a reference value, we have considered the best misclassification error registered, on the entire MAGIC dataset, over all the experiments performed (green dashed line). The cumulated costs – which are actual and not the nominal $c_1$ and $c_2$ used in the acquisition function – are also averaged on the 10 independent runs.

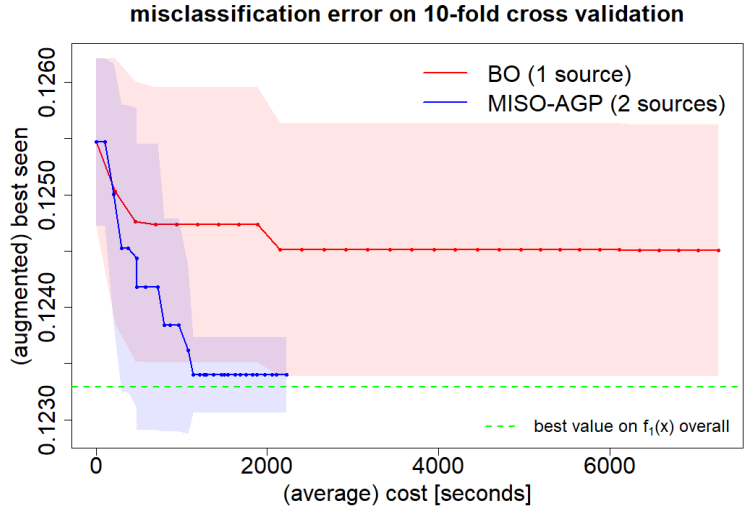

**Fig. 4.** HPO of C-SVC on the MAGIC dataset. Comparison between traditional BO based HPO and MISO-AGP on two information sources. Results refer to 10 independent runs.

The MISO-AGP approach proved to be both more effective and efficient than traditional BO: the identified hyperparameters configurations are associated to a lower misclassification error, and within less than $1/3$ of the time required by BO. On average, 60% of the function evaluations are performed on the cheaper source. Thus, MISO-AGP had intelligently exploited the cheaper information source, thanks to the proposed AGP, leading to an energy-efficient and green HPO task.

## 6    Conclusions

The GP framework can be extended to deal with multiple information sources. Relations among sources are captured by a simplified and computationally cheap discrepancy measure, which enables a sparsification strategy used to select "reliable" evaluations to fit the proposed AGP. The MISO-AGP has been empirically been shown to solve a real HPO task effectively while reducing significantly computational time and consequently energy usage.

## Acknowledgements

We greatly acknowledge the DEMS Data Science Lab, Department of Economics Management and Statistics (DEMS), for supporting this work by providing computational resources.

## Compliance with Ethical Standards

Ethical approval: This article does not contain any studies with human participants or animals performed by any of the authors.